\documentclass[dvipsnames,format=sigconf]{acmart}
\usepackage{graphicx}
\usepackage{amsmath}
\usepackage{array}
\usepackage{url}
\usepackage{amsfonts}
\newcommand{\mc}[1]{\mathcal{#1}}
\usepackage{algorithm}
\usepackage[noend]{algpseudocode}
\usepackage{verbatim}
\usepackage{multirow}
\usepackage{makecell}
\usepackage{mathtools}
\usepackage{tikz}
\usetikzlibrary{arrows.meta, positioning}
\makeatletter
\def\BState{\State\hskip-\ALG@thistlm}
\makeatother
\def\vec#1{\mathchoice{\mbox{\boldmath$\displaystyle#1$}}
{\mbox{\boldmath$\textstyle#1$}}
{\mbox{\boldmath$\scriptstyle#1$}}
{\mbox{\boldmath$\scriptscriptstyle#1$}}}

\DeclarePairedDelimiter\abs{\lvert}{\rvert}%
\AtBeginDocument{%
  }

\copyrightyear{2025}
\acmYear{2025}
\setcopyright{cc}
\setcctype{by}
\acmConference[GECCO '25 Companion]{Genetic and Evolutionary Computation Conference}{July 14--18, 2025}{Malaga, Spain}
\acmBooktitle{Genetic and Evolutionary Computation Conference (GECCO '25 Companion), July 14--18, 2025, Malaga, Spain}
\acmDOI{10.1145/3712255.3734293}
\acmISBN{979-8-4007-1464-1/2025/07}

\title[The Pitfalls and Potentials of Adding Gene-invariance to Optimal Mixing]{The Pitfalls and Potentials of\linebreak Adding Gene-invariance to Optimal Mixing}

\begin{document}


\author{Anton Bouter}
\affiliation{%
  \institution{Centrum Wiskunde \& Informatica}
  \city{Amsterdam}
  \country{The Netherlands}
}
\email{Anton.Bouter@cwi.nl}

\author{Dirk Thierens}
\affiliation{%
  \institution{Utrecht University}
  \city{Utrecht}
  \country{The Netherlands}}
\email{D.Thierens@uu.nl}

\author{Peter A.N. Bosman}
\affiliation{%
  \institution{Centrum Wiskunde \& Informatica}
  \city{Amsterdam}
  \country{The Netherlands}}
\email{Peter.Bosman@cwi.nl}


\begin{abstract}
Optimal Mixing (OM) is a variation operator that integrates local search with genetic recombination. EAs with OM are capable of state-of-the-art optimization in discrete spaces, offering significant advantages over classic recombination-based EAs. This success is partly due to high selection pressure that drives rapid convergence. However, this can also negatively impact population diversity, complicating the solving of hierarchical problems, which feature multiple layers of complexity. While there have been attempts to address this issue, these solutions are often complicated and prone to bias. To overcome this, we propose a solution inspired by the Gene Invariant Genetic Algorithm (GIGA), which preserves gene frequencies in the population throughout the process. This technique is tailored to and integrated with the Gene-pool Optimal Mixing Evolutionary Algorithm (GOMEA), resulting in GI-GOMEA. The simple, yet elegant changes are found to have striking potential: GI-GOMEA outperforms GOMEA on a range of well-known problems, even when these problems are adjusted for pitfalls - biases in much-used benchmark problems that can be easily exploited by maintaining gene invariance. Perhaps even more notably, GI-GOMEA is also found to be effective at solving hierarchical problems, including newly introduced asymmetric hierarchical trap functions.
\end{abstract}

\ccsdesc[500]{Theory of computation~Evolutionary algorithms}



\maketitle

\title[The Pitfalls and Potentials of Adding Gene-invariance to Optimal Mixing]{The Pitfalls and Potentials of Adding Gene-invariance to Optimal Mixing}

\section{Introduction}

Performing competent optimization with Evolutionary Algorithms (EAs) requires a delicate balance between two factors: search bias exploitation and diversity maintenance. On the one hand, to be as efficient as possible, we want a strong bias that fits very well with the structure of the problem being solved, so that we can use high selection pressure and exploit structural properties of our problem quickly and reliably. On the other hand, because we may not always have the perfect bias in our algorithm, we want to maintain sufficient diversity so that we avoid the risk of our EA essentially becoming a pure local optimizer that may get stuck in a local optimum because it is not performing absolutely perfect variation for our problem. Moreover, problems may have intricate, multi-layered, structure, causing our search bias to focus at first only on part of a problem. Without sufficient diversity maintenance, our search may come to a premature end on part of a problem, with a population that is fully converged already, making it impossible to still explore other parts of the problem that only appear as important later during the search process. The latter can happen in, e.g., problems with exponentially scaled substructures~\cite{thierens1998domino}, or notoriously, in hierarchical problems where the problem encompasses multiple layers, each defined in terms of multiple smaller sets of variables, such as the Hierarchical If and Only If (HIFF)~\cite{watson2000symbiotic,watson2003computational,pelikan2001escaping,thierens2013hierarchical} and the Hierarchical Concatenated Deceptive Trap (HTrap)~\cite{pelikan2000hierarchical,pelikan2001escaping} problems.
On such problems, evolution may force out optimal building blocks in favor of sub-optimal building blocks at lower levels, because the latter are easier to find, yet equally good at the lower levels.

In recent years, EAs that make use of variation operators based on the concept of Optimal Mixing (OM), have been shown to be capable of achieving state-of-the-art results on various benchmark problems as well as real-world problems, in various domains, including the binary (discrete) domain that we consider in this paper~\cite{thierens2011optimal,dushatskiy2024parameterless, goldman2014parameter,hsu2015optimization,chen2017two,przewozniczek2020empirical,przewozniczek2021direct}. Operators based on OM have high selection pressure because (small) parts of solutions are changed in variation, and subsequently accepted only if they do not decrease the fitness of the solution. When combined with a good model of which parts of solutions should be varied together, such Model-Based EAs (MBEAs) can be highly effective on problems that have a certain degree of decomposability. Achieving such models can be done either through leveraging problem-specific knowledge, or by using, e.g., Linkage Learning (LL) techniques~\cite{thierens2010linkage,chen2017two,przewozniczek2020empirical,przewozniczek2021direct,dushatskiy2021novel,przewozniczek2024direct}, which aim to discover important dependencies between problem variables, implying that these variables should (frequently) be varied \emph{jointly}. However, as outlined above, especially such EAs are at risk of premature convergence as they almost always lack a mechanism for preserving genetic material, which can cause significant portions of the population to lose valuable solution components that may be needed at a later stage during search to improve solutions further.

Various strategies have been proposed to address the issue of diversity maintenance in EAs in general, including fitness sharing \cite{holland1975adaptation,goldberg1987genetic}, crowding \cite{de1975analysis}, and other niching methods \cite{li2016seeking}, all of which aim to preserve population diversity by discouraging the dominance of a single solution.
However, these strategies often come at the cost of computational efficiency and/or additional parameters to be set, and may not be suitable for problems with complex, high-dimensional search spaces.
Gene invariance, first introduced in the Gene-Invariant Genetic Algorithm (GIGA) \cite{culberson1992genetic}, offers a more direct, and also elegant, solution to this problem by ensuring that during variation and selection, no solution component is ever discarded from the population, thereby preserving the genetic material that could contribute to better solutions perhaps at a later stage.

In this paper, we study how adding the principle of Gene Invariance (GI) may enhance OM, aimed to counteract its rapid diversity loss while not interfering with its highly efficient local-search nature of directly accepting improvements. To study this potential synergy between GI and OM, we consider the more generic framework of the Gene-pool Optimal Mixing Evolutionary Algorithm (GOMEA) \cite{dushatskiy2024parameterless,thierens2011optimal}.
In doing so, we create a new variant of GOMEA that we refer to as the Gene-Invariant GOMEA (GI-GOMEA).

In this paper, we describe the inner workings of GI-GOMEA, and experimentally test its performance using different types of linkage models on a number of well-known benchmark problems with different characteristics, ranging from deception, to hierarchy, to NP hardness.
Moreover, we introduce a number of new problems to account for biases that are found to be exploited by GI-GOMEA in some of the existing benchmark problems.
The results are compared to the original GOMEA to analyze the effect of gene invariance.


\section{Background}
\subsection{GOMEA}
The Gene-pool Optimal Mixing Evolutionary Algorithm (GOMEA) is a family of MBEAs that is considered to be among the state of the art for discrete optimization (and other problem domains)~\cite{dushatskiy2024parameterless,przewozniczek2024direct}. Different discrete GOMEA variants exist, ranging from using standard population vectors, to pyramid-structured populations, to kernel-based and multi-objective variants (see, e.g.,~\cite{thierens2011optimal,guijt2022solving,dushatskiy2024parameterless}). Each has different properties that enable good results on different types of problems, but they all share being able to leverage some form of problem decomposability by use of a linkage model that consists of multiple linkage sets, each of which identifies groups of variables that should be considered jointly during variation.

A key feature of GOMEA is the Gene-pool Optimal Mixing (GOM) variation operator, which applies variation to each solution in the population using linkage sets, as stored in the linkage model.
Specifically, for each solution in the population, a clone of that solution is made, and for each linkage set in the linkage model, a donor is randomly selected from the population. The linkage set is then used as a crossover mask, copying the genes identified by the respective linkage set from the donor to the parent.
Much like in local search, this variation step is only accepted if it does not reduce the fitness of the parent, resulting in a high selection pressure.

\subsection{Linkage Models}
A linkage model is generally described using a Family of Subsets (FOS), where each element in the FOS, also called a linkage set, denotes a subset of variables that is considered to be dependent~\cite{thierens2010linkage}.

Mostly, linkage models are defined using pairwise dependencies that can be stored in what is often called a Dependency Structure Matrix (DSM)~\cite{hsu2015optimization,helmi2012linkage,przewozniczek2024direct}. Each element in this symmetric matrix specifies the strength of a dependence between a pair of variables.

Traditionally, building upon techniques first introduced with Estimation-of-Distribution Algorithms such as ECGA~\cite{harik2006linkage} and BOA~\cite{pelikan2005bayesian}, the DSM is learned by leveraging statistical relations as observed within the population. The idea is that selection makes dependencies stand out, such that statistical means such as mutual information can be used to identify these dependencies.

More recently, fitness-based approaches have been introduced. While these require additional evaluations, an advantage is that in cases where statistical approaches do not work well, fitness-based approaches can correctly detect dependencies~\cite{przewozniczek2024direct,przewozniczek2021direct,dushatskiy2021novel}.

Alternatively, if some amount of domain knowledge is available, one can predetermine linkage models, or DSMs upon which to build linkage models. It was however previously found that it is not necessarily the case that using for linkage sets those subsets of variables for which it is known that a non-linear subfunction exists in the problem formulation, leads to the best performance. In fact, on overlapping NK landscapes, such an approach was vastly inferior to using mutual information to detect pairwise dependencies and then building a linkage tree (described below) based on the DSM.

While a linkage model can be derived in various ways, including problem-specifically by hand, if a DSM is built, different types of linkage models can be derived from it using various techniques~\cite{thierens2010linkage,hsu2015optimization,bosman2012linkage,ngai2022improving,przewozniczek2024direct,thierens2012predetermined}.
The most frequently-used, and arguably most generally useful, linkage model is the Linkage Tree (LT) \cite{thierens2010linkage}, which is a hierarchical model that captures a wide range of multivariable dependencies, ranging from univariate (i.e., no dependencies) to dependencies between very large sets of variables.
A linkage tree can conceptually be understood by starting from a set of $\ell$ subsets, each containing the index of 1 of the $\ell$ problem variables.
Larger linkage sets are added to the linkage model by continuously merging the pair of linkage sets with the largest pairwise Mutual Information (MI).
Each linkage set can only be merged into a larger set once, creating a tree structure with the root containing all problem variables.
This root node is never used as a crossover mask, as it would simply exchange complete solutions rather than parts of their genotype. While a direct implementation of the concept above leads to an algorithm that scales cubically in the number of variables, more efficient algorithms exist that scale quadratically~\cite{gronau2007optimal}

\section{Gene-Invariant GOMEA (GI-GOMEA)}
\label{sec:gigomea}
Pseudocode of GI-GOMEA is displayed in Algorithm \ref{alg:gi-gomea}. 
Before the main generational loop, we first initialize the population and evaluate it.
Rather than initializing the population by sampling each binary gene in each individual from a Bernoulli distribution with $p=0.5$, we use Probabilistically Complete (PC) sampling. Specifically, we generate all $0$ values for half the population and all $1$ values for the other half, and then shuffle the bits throughout the population per position.
This ensures that for each variable, the number of occurrences of zeroes and ones in the population is equal if the population size is even.
If the population size is odd, we perform the same process as above, but we sample the genes of the last individual in the population from a Bernoulli distribution with $p=0.5$ before shuffling.

To achieve gene-invariance, in GI-GOMEA, variation is performed on the population directly, i.e., no separate collection of offspring is generated as is the case in GOMEA. Specifically, at the start of each generation, a linkage model $\mc{F}$ is learned. Subsequently, a list of (individual, linkage-set) pairs is created by taking the Cartesian product of the population $\mc{P}$ and the linkage model $\mc{F}$.
Variation is then applied according to this list of pairs, going over it in a random order, where the individual $\vec{p}$ within this pair will serve as the parent solution and the linkage set $\mc{F}_j$ as a crossover mask.
A mate $\vec{m}$ for this parent is selected by randomly picking two different individuals in the population and selecting the best.
This tournament selection step between potential mates is done because a portion of the population will inherently have low(er) fitness due to the gene-invariant property of GI-GOMEA, decreasing the probability that an improvement can be found, and thereby decreasing the overall performance of GI-GOMEA, if no such selection is performed. Preliminary experiments confirmed this.

The GI-GOM variation operator, for which pseudocode is displayed in Algorithm \ref{alg:gi-gom}, is then applied to parent $\vec{p}$ and mate $\vec{m}$, using crossover mask $\mc{F}_j$.
Offspring of the parent $\vec{p}$ and mate $\vec{m}$, denoted $\vec{p}'$ and $\vec{m}'$ respectively, are initialized as their clones.
Crossover is then applied to this pair of offspring using crossover mask $\mc{F}_j$, exchanging all genes for which the index is contained in $\mc{F}_j$.
This operation is accepted, and $\vec{p}$ and $\vec{m}$ are overwritten by $\vec{p}'$ and $\vec{m}'$ respectively, in one of the following two scenarios:
\vspace*{-1mm}
\begin{enumerate}
    \item If the parent $\vec{p}$ is better than mate $\vec{m}$,
          and offspring $\vec{p}'$ does not have a lower fitness than $\vec{p}$.
    \item If the parent $\vec{p}$ is not better than mate $\vec{m}$,
          and offspring $\vec{m}'$ does not have a lower fitness than $\vec{m}$.
\end{enumerate}
\vspace*{-1mm}
In Algorithm \ref{alg:gi-gom}, the fitness of a solution $\vec{x}$ is determined by a call to $f(\vec{x})$, which performs a fitness evaluation.
Note that the two offspring solutions $\vec{p}'$ and $\vec{m}'$ are not necessarily both evaluated.
Specifically, if the variation step is rejected, the other offspring is not evaluated.
Because the variation operator accepts either the original pair of parent and mate, or the pair of offspring which had their genetic material exchanged, no genetic material is ever lost and the gene-invariant property is maintained.

\section{Experiments}
We experimentally test\footnote{A setting for GI is available in the GOMEA library: \url{https://github.com/CWI-EvolutionaryIntelligence/gomea}} and compare the performance and scalability of GI-GOMEA and GOMEA, both combined with various linkage models. Identical to GI-GOMEA, probabilistically complete sampling was also used in GOMEA upon initialization.

On the NK-S1 and MaxCut problems, 50 unique instances were used for each problem dimensionality, each optimized once per experiment of 50 runs.
These sets of 50 instances were identical for the experiments with different (variations of) algorithms.

\begin{algorithm}%
\small
\caption{GI-GOMEA}\label{alg:gi-gomea}
\begin{algorithmic}[1]
\Procedure{$\texttt{GI-GOMEA}$}{$n$}
\State $\mc{P} \gets \texttt{InitializeAndEvaluatePopulation}(n)$ \Comment{\scalebox{0.9}[1]{\emph{PC}}}
\While{$\neg \texttt{TerminationCriterionSatisfied}()$}
    \State $\mc{F} \gets \texttt{LearnLinkageModel}(\mc{P})$
    \State $\texttt{GOMItems} \gets \mc{P} \times \mc{F}$                                           \Comment{\scalebox{0.9}[1]{\emph{Cartesian product}}}
    \For{$(\vec{p},\mc{F}_j) \in \texttt{GOMItems}$}                                                \Comment{\scalebox{0.9}[1]{\emph{Random order}}}
        \State $\vec{m} \gets \texttt{BestOf2RandomIndividuals}(\mc{P}-\vec{p})$
        \State $(\vec{p},\vec{m}) \gets \texttt{GI-GOM}(\vec{p},\vec{m},\mc{F}_j)$                  \Comment{\scalebox{0.9}[1]{\emph{Replace in $\mc{P}$}}}
    \EndFor
\EndWhile
\EndProcedure
\end{algorithmic}
\end{algorithm}

\begin{algorithm}%
\small
\caption{GI-GOM}\label{alg:gi-gom}
\begin{algorithmic}[1]
\Procedure{$\texttt{GI-GOM}$}{$\vec{p},\vec{m},\mc{F}_j$}
\State $\vec{p}' \gets \vec{p}$                     \Comment{\scalebox{0.9}[1]{\emph{Copy parent}}}
\State $\vec{m}' \gets \vec{m}$                     \Comment{\scalebox{0.9}[1]{\emph{Copy mate}}}
\For{$u \in \mc{F}_j$}                              \Comment{\scalebox{0.9}[1]{\emph{Use $\mc{F}_j$ as crossover mask}}}
    \State $\vec{p}'[u] \gets \vec{m}[u]$
    \State $\vec{m}'[u] \gets \vec{p}[u]$
\EndFor
\If{$\vec{p} == \vec{p}'$}                          \Comment{\scalebox{0.9}[1]{\emph{Nothing was changed}}}
     \State \Return $(\vec{p},\vec{m})$
\EndIf
\If{$f(\vec{p}) > f(\vec{m})$}                      \Comment{\scalebox{0.9}[1]{\emph{Parent better than mate}}}
    \If{$f(\vec{p}')\! \geq\! f(\vec{p})$}              \Comment{\scalebox{0.9}[1]{\emph{And better or same after crossover}}}
        \State \Return $(\vec{p}',\vec{m}')$        \Comment{\scalebox{0.9}[1]{\emph{Accept variation (and evaluate $\vec{m}')$}}}
    \EndIf
\Else                                               \Comment{\scalebox{0.9}[1]{\emph{Mate at least as good as parent}}}
    \If{$f(\vec{m}')\! \geq\! f(\vec{m})$}              \Comment{\scalebox{0.9}[1]{\emph{And better or same after crossover}}}
        \State \Return $(\vec{p}',\vec{m}')$        \Comment{\scalebox{0.9}[1]{\emph{Accept variation (and evaluate $\vec{p}'$)}}}
    \EndIf
\EndIf
\State \Return $(\vec{p},\vec{m})$                  \Comment{\scalebox{0.9}[1]{\emph{Reject variation}}}
\EndProcedure
\end{algorithmic}
\end{algorithm}

\subsection{Linkage Models}
In this paper, we always learn an LT from the DSM. The LT offers a generic way to represent dependencies at different scales, while the content of the LT, i.e., what is in the subsets in the tree, still depends on the DSM. For the DSM content, we consider 3 alternatives.

\subsubsection{Mutual Information (MI)}
The most commonly adopted approach to learning dependencies using population statistics, is to use Mutual Information (MI) estimated from the population~\cite{thierens2011optimal}.
In GI-GOMEA, we estimate the MI from the best 50\% of the population however, because the gene-invariant property will inherently cause low-quality individuals to remain in the population.

\subsubsection{Distance-based Dependence Estimate (dDSM)}
In real-world scenarios some general notion of which variables depend on each other may be available, but no exact dependence relations are known. A predetermined DSM can then still be used. To do so, we assume that according to the available notion of dependence, the variables in the problem can be re-ordered so that dependent variables are tightly linked, i.e., encoded close to each other. For this case, we set the dependency strength between a pair of variables $\vec{x}_i$ and $\vec{x}_j$, i.e., with index $i$ and $j$ in the encoding, to a distance-based relation: $\ell - \abs{i-j}$, where $\ell$ is the total number of problem variables.

\subsubsection{Weighted VIG Dependence (wvigDSM)}
More specific knowledge about the optimization problem may be available, or could have a priori been learned using a different linkage learning approach, e.g., fitness-based linkage learning~\cite{yang2020efficient,przewozniczek2020empirical,dushatskiy2021novel,przewozniczek2024direct}.
In this case, we assume that linkage information can be provided in the form of a weighted Variable Interaction Graph (VIG), indicating which pairs of variables have a direct dependency and a measure of the strength of this dependency.
For instance, in the well-known MaxCut problem \cite{karp1972reducibility}, the graph defining the problem instance is an exact representation of such a weighted VIG.

\subsection{Benchmark Problems}
We consider various benchmark problems that are commonly used to test the performance of MBEAs. We also consider two variants of an NP-hard problem: Maximum Cut (MaxCut). We subdivide these problems according to whether or not they are symmetric.

The unitation function $u(\vec{x})$ is used in various definitions, which is defined as the number ones in the vector $\vec{x}$.
In the hierarchical trap functions, the symbol '-' may be part of $\vec{x}$, in which case $u(\vec{x})$ is undefined.
Any fitness contribution by $\vec{x}$ will then be equal to 0.

\subsubsection{Symmetric Benchmark Problems}
\label{subsec:symmexp}
All problems in this section have a specific symmetry property in that the global optimum is the complement of the local optimum/deceptive attractor.

\paragraph{Concatenated Deceptive Trap Functions}
The well-known concatenated deceptive trap function with subfunction size $k$~\cite{deb1993analyzing}, here referred to as TrapK, is perfectly separable into non-overlapping blocks of size $k$. For each block, the subfunction defined in Eq.~\ref{eq:trapk} is used. The final problem is to maximize the sum of these subfunctions. For each block, a very high probability of not being disrupted during crossover is required for an EA to be successful.
For this problem, the DSM in the wvigDSM approach is filled with an arbitrary constant value for each pair of variables within the same block, and 0 otherwise.

\vspace*{-3mm}
\begin{align}
\label{eq:trapk}
g_{\texttt{TrapK}}(\vec{x},k) &= \begin{cases}~1 & \texttt{if } u(x)=k \\
                                ~\frac{k-u(\vec{x})-1}{k} & \texttt{otherwise}
                                \end{cases}
\end{align}

\paragraph{Hierarchical Trap Functions}
The hierarchical trap (HTrapK) problem~\cite{pelikan2000hierarchical,pelikan2001escaping} with trap size $k$ can best be conceptualized using a tree.
In this tree, the lowest level consists of the genes of the solution $\vec{x}$.
These genes are grouped into sets of $k$ consecutive bits, which map to one of the symbols 0, 1, or '-', in the next level of the tree.
If any such group of genes consists of \emph{only} zeroes or ones, it maps to a 0 or 1, respectively.
Otherwise, it maps to the symbol '-'.
The same mapping is applied at each level in the tree for $k$ consecutive symbols.
Each group of $k$ symbols in each level of the tree contributes to the fitness of $\vec{x}$ by the amount $g_{\texttt{HTrapK}}$ (Eq.~\ref{eq:htrapk-eq1}), which are essentially the same subfunctions as in the concatenated deceptive trap functions, but here the suboptimum and the optimum have the same fitness, except at the top level.
Each subfunction is weighted by $k^h$ where $h$ is the height of the tree level, starting from 1 at the lowest level.
In Eq.~\ref{eq:htrapk-eq1}, $w(h)$ is 0.9 at the top-most level and 1 otherwise.
When using a linkage tree based on wvigDSM, we fill the DSM such that the dependency between a pair of variables is equal to $k^{(q-1)}$ where $q$ is the number of levels in the tree in which the pair of variables is input to the same subfunction.

\vspace*{-3mm}
\begin{align}
\label{eq:htrapk-eq1}
g_{\texttt{HTrapK}}(\vec{x},k,h) &= 
    \begin{cases}
        ~0 & \texttt{if '-' in }\vec{x}\\
        ~1 & \texttt{elif } u(x)=k \\
        ~w(h)\frac{k-u(\vec{x})-1}{k-1} & \texttt{otherwise}
    \end{cases}
\end{align}

\paragraph{Hierarchical If and Only If (HIFF)}
The HIFF problem~\cite{watson2000symbiotic,watson2003computational,pelikan2001escaping,thierens2013hierarchical} is a hierarchical problem that can be visualized using a binary tree structure.
The leaves of the tree consist of the genes of a solution $\vec{x}$, and adjacent groups of nodes are merged until the root node that includes all genes.
Each node of size $k$ in this tree contributes to the fitness of $\vec{x}$ if all genes described by the node, i.e., their values in $\vec{x}$, are identical.
If this is the case, the node contributes $k$ to the fitness value, otherwise 0.
Hence, the optimum of this function is either all ones or all zeroes.
For this problem, the DSM in the wvigDSM approach is filled such that for each pair of variables the DSM entry is equal to $k^{(q-1)}$ where $q$ is the number of levels in the tree in which the pair of variables is input to the same subfunction.

\subsubsection{Asymmetric Benchmark Problems}
\label{subsec:asymmexp}
We additionally consider problems that are not symmetric in their optimum and suboptima.

\paragraph{Concatenated Bimodal Deceptive Trap Functions}
This problem (BimTrapK) is a sum of multiple concatenated bimodal deceptive trap functions, defined in Eq.~\ref{eq:bimk}. 
BimTrapK contains a subfunction of size $k$ that has two optima: either when the genes are all zeroes, or when they are all ones.
Moreover, each configuration with an equal number of zeroes and ones is locally optimal, resulting in a very large number of local optima.
For this problem, the DSM in the wvigDSM approach is the same as for the TrapK problem.

\vspace*{-3mm}
\begin{align}
\label{eq:bimk}
g_{\texttt{BimTrapK}}(\vec{x},k) &\!=\! \begin{cases} 
                                ~1 & \texttt{if } \!u(x)\!=\!0\! \texttt{ or } \!u(x)\!=\!k\!\\
                                 ~\frac{k\!-\!\abs{2u(\vec{x})\!-\!k}\!-\!2}{k} & \texttt{otherwise}
                            \end{cases}
\end{align}

\paragraph{Concatenated Asymmetric Deceptive Trap Functions}
The asymmetric deceptive trap problem with trap size $k$ (AsymTrapK), is a variation of the concatenated deceptive trap problem for which the global optimum and the deceptive attractor are not each other's complement.
Its subfunctions, described in Eq.~\ref{eq:asymtrapk-subf}, are concatenated $\ell/k$ times.
The global optimum of each trap now consists of one 0 in the first location, and all ones in the remaining locations.
Moreover, we set the fitness of the former optimum (all ones) to 0 by increasing the numerator and denominator of the function by 1.
For this problem, the DSM in the wvigDSM approach is a block matrix with blocks of size $k$ where each block is filled with an arbitrary positive constant, and 0 otherwise.

\vspace*{-3mm}
\begin{align}
\label{eq:asymtrapk-subf}
g_{\texttt{AsymTrapK}}(\vec{x},k) &\!=\! \begin{cases} 
                                ~1 & \texttt{if } \!u(x)\!=\!k\!-\!1\!\texttt{ and } \!\vec{x}[0]\!=\!0\!\\
                                ~\frac{k-u(\vec{x})}{k+1} & \texttt{otherwise}
                            \end{cases}
\end{align}

\paragraph{Asymmetric Hierarchical Trap Functions}
The asymmetric variant of the HTrapK problem (AsymHTrapK) has the same hierarchical structure as the HTrapK problem, but has a different mapping of symbols within the tree, and different fitness contributions.
Similar to the AsymTrapK problem, the optimal configuration of any block no longer consists of all ones, but has a 0 in the first location.
The mapping ensures that a block of $k$ such optimal blocks maps to one optimal block in the higher level, which is either a 0 or 1 depending on its location.
Hence, one optimal block maps to either a 0 or a 1 depending on its location.
A block of $k$ zeroes still maps to a 0, and any remaining configurations map to the symbol '-'.
To ensure that the optimum consists of all building blocks of $[01\dots1]$, rather than all zeroes in the first $\ell/k$ blocks, there is no longer a distinction between the fitness contributions of blocks at different levels.
At each level, the fitness contribution of any block is given by Eq.~\ref{eq:asymhtrapk-eq1}, again weighted by $k^h$ with $h$ the height of the level.
We consider the AsymHTrap4 problem, because the problem is no longer deceptive when an asymmetric trap function size of 3 is used.
When using a linkage tree based on wvigDSM, we fill the DSM the same way as for the hierarchical trap functions.

\vspace*{-3mm}
\begin{small}
\begin{align}
\label{eq:asymhtrapk-eq1}
g_{\texttt{AsymHTrapK}}(\vec{x},k) &= 
    \begin{cases} 
        ~0 & \texttt{if '-' in }\vec{x}\\
        ~1 & \texttt{elif } \!u(x)\!=\!k\!-\!1\!\texttt{ and } \!\vec{x}[0]\!=\!0\!\\
        ~0.9\frac{k-u(\vec{x})}{k} & \texttt{otherwise}
    \end{cases}
\end{align}
\end{small}

\paragraph{Consecutively Overlapping NK Landscapes}
We consider the NK-Landscapes~\cite{kauffman1993origins,kauffman1987towards} function with consecutively encoded subfunctions of size 5 and a shift of 1, referred to as NK-S1~\cite{pelikan2009performance}.
This benchmark problem consists of $\ell-k+1$ subfunctions of groups of $k$ consecutive genes, where the starting index of each subfunction shifts by one.
Weights of all $2^k$ configurations in each subfunction are generated uniformly at random between 0 and $10^6$.
Cells in the DSM for the wvigDSM model are filled with the number of subfunctions that the respective pair of variables has in common.

\paragraph{Maximum Cut}
We consider two types of instances of the NP-hard Maximum Cut (MaxCut) problem~\cite{karp1972reducibility}. The objective is to assign each node in a weighed graph to one of two sets, such that the sum of weights of edges connecting any two nodes in different sets, is maximized. Each node is encoded with one bit.

The instances we consider are both fully connected.
The first type of instance, referred to as MaxCut-Full, has weights randomly generated from a Beta distribution with parameters $\alpha=100$ and $\beta=1$, and scaled to the range $[1:5]$.
Instances of the second type, referred to as MaxCut-Geo, were generated by randomly distributing $\ell$ points, representing vertices, in a $1000\times1000$ square and using the floored Euclidean distance between each pair of points as the weight of the edge between the vertices that they represent. These instances have a bit more structure as the edge weights are distances that adhere to the triangle inequality.

For this problem, the DSM in the wvigDSM approach is filled with the absolute values of the edge weights.
Though the weights were positive in all our instances, taking the absolute value is recommended in a more general setting. Moreover, we do not consider the dDSM approach for MaxCut as it is not a priori clear how to ensure tight linkage in a linear encoding of a MaxCut instance.

\subsection{Preliminary Analysis - The Pitfalls}
\label{subsec:prelim}
\begin{figure*}[htbp]
\scalebox{0.94}
{
\begin{tabular}{cccc}
    \textbf{\makecell{81-D HTrap3}} &
    \textbf{\makecell{32-D BimTrap8}} &
    \textbf{\makecell{40-D BimTrap10}} \\
    \includegraphics[width=0.33\linewidth]{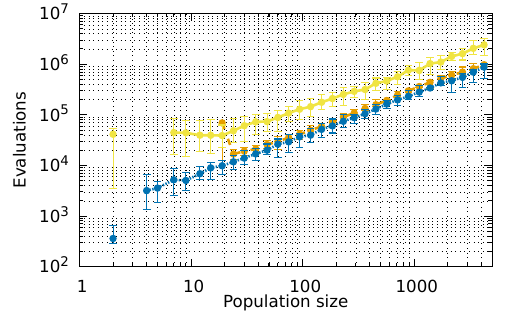} &
    \includegraphics[width=0.33\linewidth]{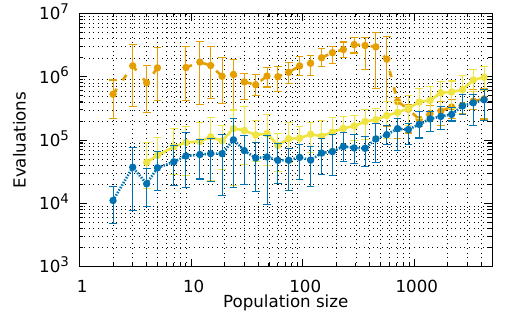} &
    \includegraphics[width=0.33\linewidth]{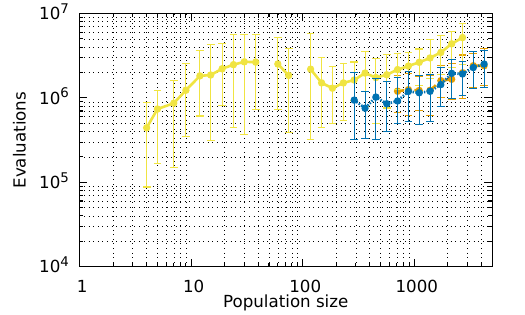} \\
    \multicolumn{3}{c}{\includegraphics[width=0.6\linewidth]{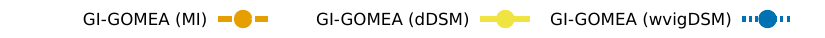}} \\
\end{tabular}
}
\vspace*{-5mm}
\caption{Median and interdecile ranges (50 runs; at least 49 successful; budget $10^8$ function evaluations) of the number of evaluations to find the optimum of various symmetric benchmark problems using exponentially increasing population sizes between 2 and 5000 for variants of GI-GOMEA. Note differences in scale between plots.}
\label{fig:popsize-snakes}
\end{figure*}

In preliminary experiments, we observed unusual behavior in symmetric problems. In Figure~\ref{fig:popsize-snakes}, we show the required function evaluations to solve the hierarchical trap functions for $\ell=81$. The concatenated trap and HIFF functions behaved similarly. When provided the true VIG information in the DSM, all these problems can be solved for a population size as small as $n=2$. The reason is, that even if the algorithm gets deceived and pulled toward the suboptimum for all subfunctions, because it always keeps the complement genes in the other solution, in the end, the optimum is still found in the other solution. Similarly unexpected is the fact that for population size $n=3$, the problem does not get solved, even with wvigDSM. The reason for this is that there are now 2 other solutions instead of 1 across which the missing information can be spread, so the optimum is not necessarily found when the suboptimum is found. Certainly, this is to be considered a potentially unfair bias, calling for the need to have better benchmark problems, or to be aware of this situation when making comparisons between algorithms. Clearly, GI-GOMEA does not behave the same when it comes to population sizing as typical selectorecombinative GAs do, or even GOMEA does. When using MI, the problem cannot be solved for very small population sizes, since there is not enough statistical information in the population in any generation to deduce the required structural information. When using dDSM, the boundaries of the subfunctions are not explicitly encoded, but there is a bias that minimizes breaking of building blocks for subfunctions higher up in the linkage tree, where the larger linkage sets are, which contain tightly coded indices, just as the problem does. As such, there is an overhead compared to MI and wvigDSM which can efficiently find or are given the building block boundaries, but all problems can be solved, again even for $n=2$.

However, while forewarned about symmetric problems in the original paper on gene invariance in genetic algorithms~\cite{culberson1992genetic}, unusual behavior is also observed on the bimodal trap problem (BimTrapK)~\cite{deb1993multimodal} with trap size $k=10$. This problem is not symmetric in its optima or suboptima, but it has many symmetries in terms of unitation, which is the root cause for the observed unusual behavior. Figure \ref{fig:popsize-snakes} shows, that the bimodal trap functions can still be solved to optimality by GI-GOMEA with very small population sizes and the wvigDSM, but not for $k=10$. The reason for this is particular to the use of linkage learning and is beyond the original paper on gene invariance.
Specifically, because each dependency in a block of the wvigDSM has an equal weight, balanced linkage trees get built. 
In the case of $k=10$, all sets of size 1 are merged into 2, and all sets of 2 into 4.
Blocks of 6 and 10 can also be represented (by merging a block of 4 and 2 and a block of 6 and 4, respectively), but a block of 5 bits does not appear in the LT. Now, as suboptima are found for bimodal trap subfunctions, half the bits are 0 and half are 1. Other solutions also have such combinations. However, one requires crossing over exactly those bits that are all the same within one subfunction to achieve an improvement, i.e., $4$ bits for $k=8$, and $5$ bits for $k=10$. For $k=8$, linkage sets of size 4 are indeed in the LT. Moreover, because GI-GOM accepts changes even if the fitness remains the same, it will keep stepping through local optima until at some point it happens to have the right mask of 4 bits and the right donor to crossover the block of complementary bits to reach the optimal block. For $k=10$ this is thus not possible, given a perfectly balanced LT. For dDSM, the tree is also balanced, but not necessarily in even linkage sets. As such, it can solve the problem for $k=10$ with much smaller population sizes, employing the same “wait”-and-“move across the suboptimal plateau” approach. Again, even though the bimodal trap functions are not symmetric in their optimum and suboptimum, there is an unfairly exploitable bias that GI-GOMEA can employ, making these problems not a good basis for comparison by themselves. This is the reason why we defined the asymmetric benchmark functions and why we have performed our final experiments, in which we ascertain the scalability when adding GI to GOMEA, on both the classic symmetric problems and our newly defined asymmetric problems.

\begin{figure*}[htbp]
\scalebox{0.94}
{
\begin{tabular}{cccc}
    \centering
    \textbf{\makecell{Trap5}} &
    \textbf{\makecell{HTrap3}} &
    \textbf{\makecell{HIFF}} \\
    \includegraphics[width=0.33\linewidth]{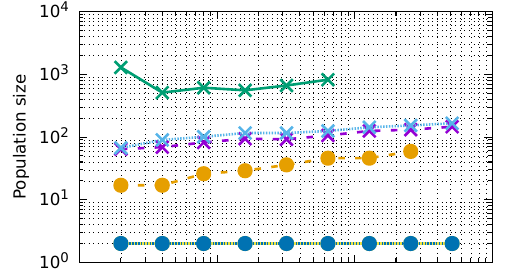} &
    \includegraphics[width=0.33\linewidth]{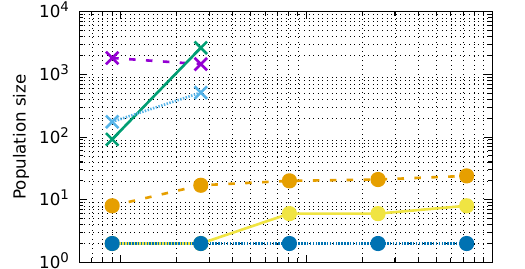} &
    \includegraphics[width=0.33\linewidth]{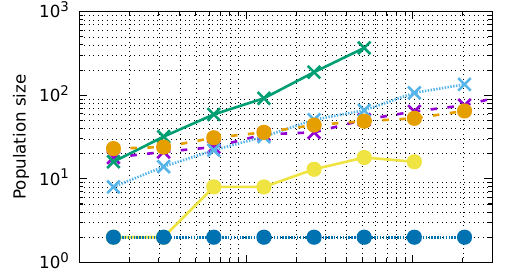} \\
    \includegraphics[width=0.33\linewidth]{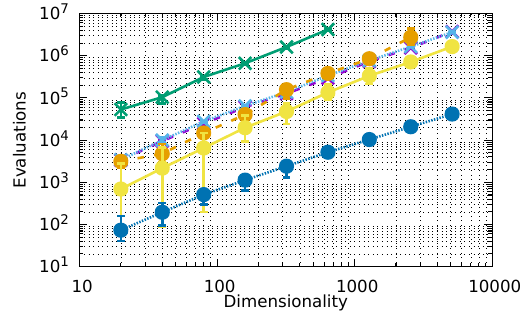} &
    \includegraphics[width=0.33\linewidth]{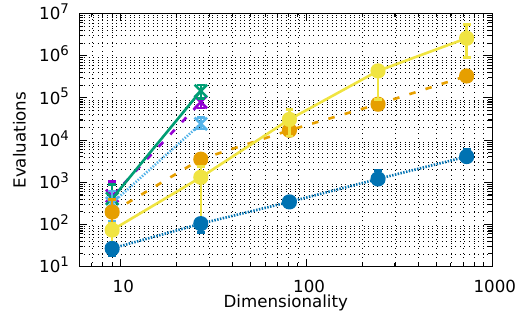} &
    \includegraphics[width=0.33\linewidth]{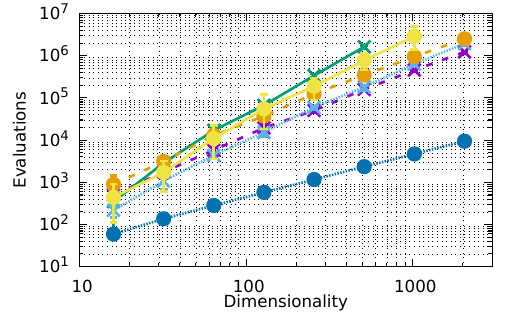} \\
    \multicolumn{3}{c}{\includegraphics[width=0.7\linewidth]{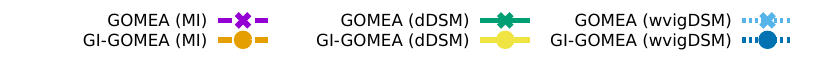}} \\
\end{tabular}
}
\vspace*{-5mm}
\caption{Top row: Estimated optimal population sizes (see Section \ref{subsec:optpop}). Bottom row: Corrected number of function evaluations and interdecile ranges of function evaluations of successful runs (50 independent runs) to reach the optimum using the given optimal population size. Note differences in scale between plots.}
\label{fig:results-symm}
\vspace*{-5mm}
\end{figure*}


\renewcommand{\arraystretch}{0.8}
\begin{figure*}[htbp]
\scalebox{0.94}
{
\begin{tabular}{cccc}
    \centering
    \textbf{\makecell{BimTrap10}} &
    \textbf{\makecell{AsymTrap5}} &
    \textbf{\makecell{AsymHTrap4}} \\
    \includegraphics[width=0.33\linewidth]{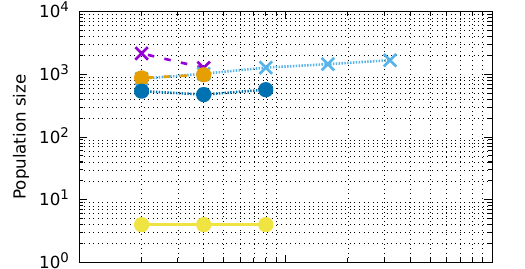} &
    \includegraphics[width=0.33\linewidth]{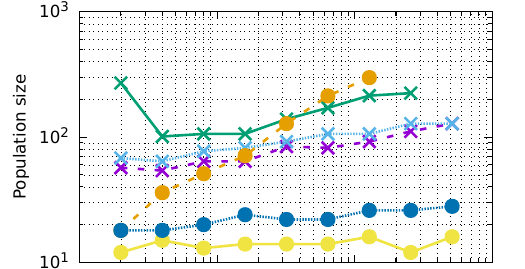} &
    \includegraphics[width=0.33\linewidth]{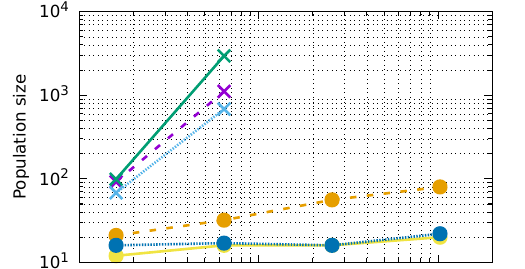} \\
    \includegraphics[width=0.33\linewidth]{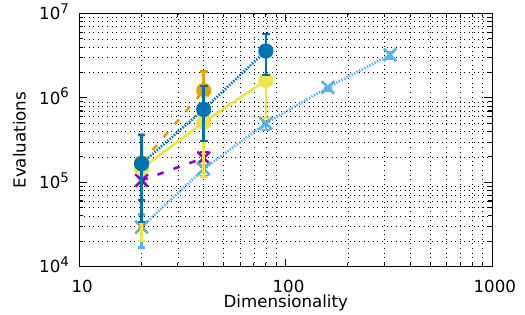} &
    \includegraphics[width=0.33\linewidth]{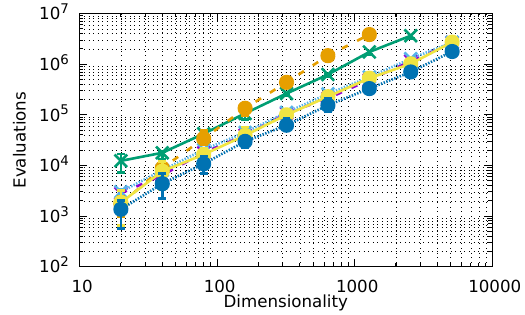} &
    \includegraphics[width=0.33\linewidth]{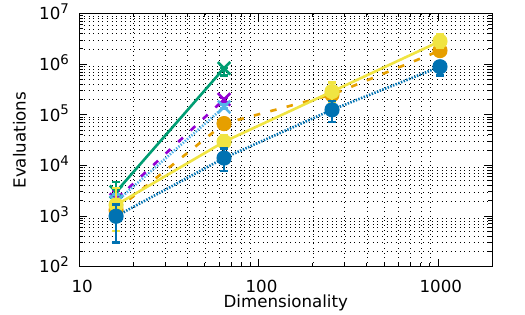} \\
    \textbf{\makecell{NK-S1}} &
    \textbf{\makecell{MaxCut-Full}} &
    \textbf{\makecell{MaxCut-Geo}} \\
    \includegraphics[width=0.33\linewidth]{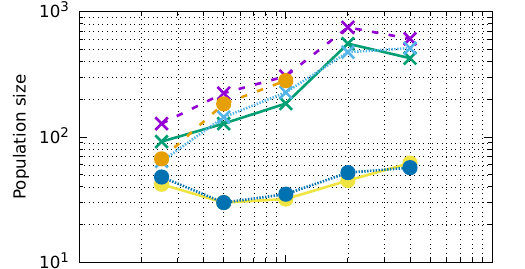} &
    \includegraphics[width=0.33\linewidth]{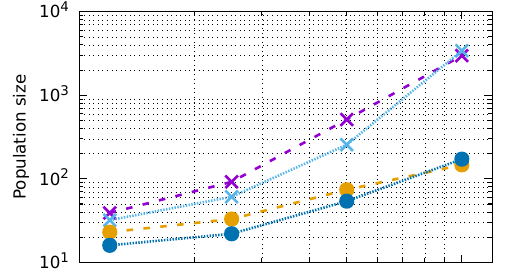} &
    \includegraphics[width=0.33\linewidth]{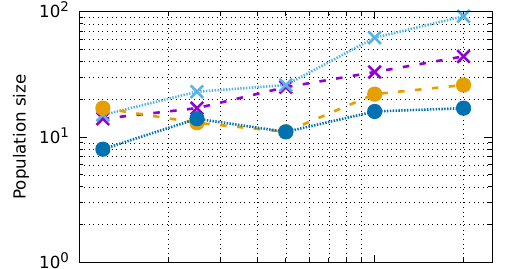} \\
    \includegraphics[width=0.33\linewidth]{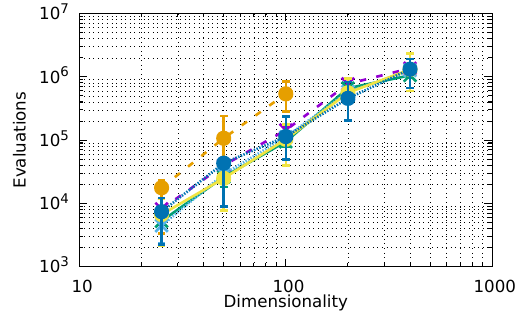} &
    \includegraphics[width=0.33\linewidth]{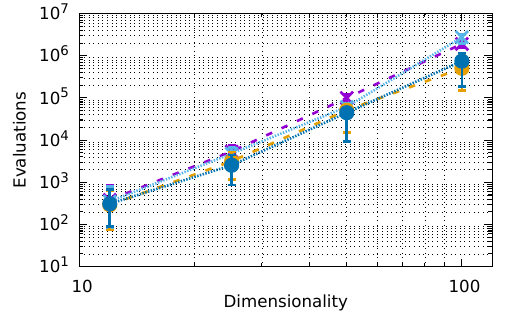} &
    \includegraphics[width=0.33\linewidth]{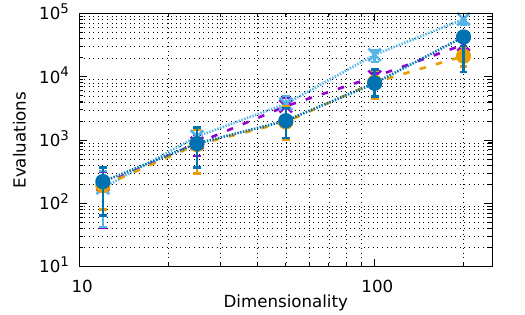} \\
    \multicolumn{3}{c}{\includegraphics[width=0.7\linewidth]{plots/key.pdf}} \\
\end{tabular}
}
\vspace*{-5mm}
\caption{First and third rows: Estimated optimal population sizes (see Section \ref{subsec:optpop}). Second and fourth rows: Corrected number of function evaluations and interdecile ranges of function evaluations of successful runs (50 independent runs) to reach the optimum using the given optimal population size. Note differences in scale between plots.}
\label{fig:results-asymm}
\end{figure*}

\vspace*{-3mm}
\subsection{Scalability Analysis - The Potentials}
\subsubsection{Setup}
\label{subsec:optpop}
In this section we aim to assess the theoretical lower bound that best showcases the potential of the considered algorithms through a scalability analysis. To do so, we aim to find the population size that minimizes the average number of function evaluations (of successful runs) to reach the optimum with a success rate of 49 out of 50 runs within a budget of $10^7$ function evaluations, for various problem dimensionalities $\ell$.

We used the following procedure to estimate the optimal population size.
Due to the sometimes unusual behavior of the required number of function evaluations (see Section~\ref{subsec:prelim} and Figure~\ref{fig:popsize-snakes}), we make a first estimate of the optimal population size  (denoted $n^{\texttt{est}}$) for an EA by running that EA for $n \in [2,4,8,\dots,8196]$ and choosing the one for which the smallest average number of required evaluations is observed.
If the required success rate is not achieved for any of the trial population sizes, the experiment is considered unsuccessful.
Otherwise, we perform a bisection in the range $[0.5n^{\texttt{est}},2n^{\texttt{est}}]$ to find the final estimate of the optimal population size.
We repeat this procedure (initial estimate and bisection) $5$ times, and use the average thereof as the final estimate $\hat{n}^{\texttt{opt}}$ of the true optimal population size $n^{\texttt{opt}}$, if all five iterations were successful. Finally, we perform another 50 independent runs with $\hat{n}^{\texttt{opt}}$ to get final results.

\subsubsection{Results}
In Figure~\ref{fig:results-symm} we show all the scalability results on the symmetric problems and in Figure~\ref{fig:results-asymm} for the asymmetric problems.
Because the success rate of each EA using its $\hat{n}^{\texttt{opt}}$ can still vary slightly in the final experiments, we report the corrected average number of function evaluations~\cite{suganthan2005problem}, defined as the average number of evaluations of successful runs divided by the success rate.

\paragraph{Symmetric problems}
Following what we already observed in Section~\ref{subsec:prelim}, it becomes again immediately clear from Figure~\ref{fig:results-symm} that GI-GOMEA can exploit the symmetry present in these types of problems, in particular when an accurate linkage model is used, because it can effectively solve these problems with a population size of 2.
However, this is not always the case when a less accurate linkage model (here, dDSM) is used, which is not always successful with a population size of 2, though nevertheless always requires a smaller population size than GOMEA.
Moreover, when using MI, the required population size of GI-GOMEA is much larger so that accurate linkage learning may be performed, though it is still smaller or similar to that of GOMEA. The required number of function evaluations is however not always smaller when classical statistical linkage learning with MI performed. However, we point out that the classic notion of learning linkage only from the best solutions in a selectorecombinative EA is not valid in GI-GOMEA as it explicitly also maintains less good solutions as needed to maintain gene invariance. As such, learning linkage using statistical measures in a GI context requires revisiting. The (theoretical) potential however, given by the use of wvigDSM in GI-GOMEA is vast, but can also be called unfair or biased on the symmetric problems.

\paragraph{Asymmetric problems}
The positive impact of adding GI to OM remains, even when problems become asymmetric. Clearly, on these problems GI-GOMEA can no longer ``cheat'' and find optimal solutions with population sizes $2$ or $4$ (except for the otherwise still highly symmetric bimodal trap functions, as discussed). 

Importantly, GI-GOMEA still solves the hierarchical
asymmetric trap function for non-trivial problem dimensionalities, whereas the original GOMEA cannot (contrary to what was reported in~\cite{thierens2013hierarchical} and confirmed in~\cite{ngai2022improving}) as it converges too fast on the lower layers of the hierarchical problem. Although this is not the first MBEA capable of solving hierarchical problems~\cite{pelikan2000hierarchical,ngai2022improving}, it does so using an elegant and relatively simple adjustment of the GOM operator, whereas previous attempts required more extensive adjustments. Moreover, the most competent MBEA to solve hierarchical problems so far, DSMGA-II with preservative back mixing~\cite{ngai2022improving}, explicitly stores patterns that are equally good. This does not generalize to the case represented by our AsymHTrap function in which on every level the suboptimal attractor does not have the same fitness. We argue therefore that adding GI to GOM as proposed here, is more general.

Another observation that underscores the potential of adding GI to OM is that on both MaxCut instances, GI-GOMEA scales better than GOMEA, even though other MBEAs that added hierarchical problem solving capabilities or other MBEAs based on optimal mixing in general performed worse on this problem than GOMEA.

What furthermore stands out, is that the minimally required population size for GI-GOMEA is smaller than for GOMEA, sometimes vastly so. While in principle the most important is the required number of function evaluations, the fact that population sizes can be much smaller does hold additional potential. When tackling multi-objective optimization, or multi-structured problems in general, it is known that applying clustering techniques along the Pareto front~\cite{bosman2003balance,pelikan2005multiobjective,luong2014multi} and using linkage kernels in which each solution can mate only with a small number of nearby solutions~\cite{guijt2022solving}, enhances the problem-solving capabilities of GOMEA. In both cases, being able to get good results with very small population sizes is highly beneficial. As such, GI-GOMEA may hold even more potential than what we currently observe in the scalability graphs.

Besides the results on the bimodal trap functions with $k=10$ being biased for the dDSM, leading to extremely small population sizes, it is also interesting to see that actually GI-GOMEA does not work very well, at least not compared to GOMEA with wvigDSM. It is known that it is difficult to learn the structure of the bimodal trap functions using mutual information and that other means of learning linkage work much better here~\cite{przewozniczek2020empirical,przewozniczek2024direct}. But even when the information is available, GI-GOMEA does not scale well on the concatenated bimodal trap functions with $k=10$. It appears to scale well on the asymmetric trap functions with $k=5$, but if $k$ is enlarged for that problem, GI-GOMEA also starts to scale less well. The main reason for this, is that a drawback in GI-GOMEA is that by design it cannot let the proportion of building blocks grow in the population. If there are 2 partitions with 1 optimal building block each, this building block can only, in the best case, be passed to another individual, but it cannot be replicated in multiple individuals. Hence, variation must wait until the right parent and donor are selected that have these building blocks. This requires in expectation a number of mixing attempts that is quadratic in the population size. Such negative scaling effects are not seen in any of the other problems which are (hierarchically) overlapping. It appears that this is where GI has the most added value.

Arguably however, overlapping problems are the only problems of real interest. It can be seen from the results on the asymmetric trap functions that statistical linkage learning with MI does not work well. It would be good to move to more recent manners of linkage learning that are fitness-based~\cite{przewozniczek2024direct,przewozniczek2021direct,dushatskiy2021novel}, or to assume that VIG information can be supplied otherwise. With such approaches, it would be readily clear (from the DSM) that the bimodal trap functions and the (asymmetric) trap functions consist of fully separable subfunctions. In such a case, if these subfunctions are large, one would run GI-GOMEA on each such subfunction (in parallel and within a template solution) and combine the best results, or if the subfunctions are small, one would perform exhaustive search on each subfunction (within a template solution) and combine the best results. The latter would result in $1+2^{k}\frac{\ell}{k}$ function evaluations for subfunctions of size $k$. Adding to this that it is known that a full Walsh decomposition that would reveal all dependencies can be performed in ${\mathcal O}(\ell2^k)$ time~\cite{dushatskiy2021novel}, with for most problem empirically observed $10\ell2^k$ evaluations, additively decomposable problems would be solvable in approximately $1+2^k\ell(10+\frac{1}{k})$ function evaluations without a priori known linkage information, which corresponds to 3.3 million function evaluations for the largest instance (320 bits) solved in this paper for the concatenated bimodal trap functions and 1.7 million function evaluations for the largest asymmetric trap functions instance (5120 bits) solved in this paper, which are very close to the best found results of GOMEA and GI-GOMEA, but then leveraging the predefined wvigDSM on these respective problems.

As a final remark on these results, we note that when using wvigDSM, the results on MaxCut-Geo actually become worse than when using MI. This is an interesting observation, as it signals that there is more to successfully evolving a population of solutions than having the perfect problem decomposition information and placing that inside a linkage model. The information provided by the population itself, which indicates the actual genetic material we have to work with, may be more valuable even as it can also signal for instance on which part of the problem the EA has already converged more. A similar outperformance of the use of MI compared to wvigDSM can also be observed on the (asymmetric) trap functions where GOMEA with MI outperforms GOMEA with wvigDSM, confirming results previously already published about the use of fixed linkage structures~\cite{thierens2012predetermined}. This signals that even if we would know, or be able to construct, the perfect weighted VIG information, there is potentially still additional (linkage) value to be exploited by incorporating the information found in the evolving population, which would be an interesting avenue of future research.

\section{Discussion and Conclusions}
In this paper, for the first time, we have combined the concepts of Gene Invariance (GI) and Optimal Mixing (OM). We brought these concepts together in the modern MBEA known as GOMEA and in doing so established GI-GOMEA. Importantly, this paper was not intended to establish the best possible new variant of GOMEA or any modern MBEA that leverages OM, but rather, to assess the potential of adding GI to OM. We argue that our experiments have certainly outlined that this combination has interesting new, performance enhancing potential, but we explicitly define it to be future work to engineer the most out of this combination.

Through various experiments, including scalability experiments, we have found that GI has the potential to increase the performance of OM on the hardest problems that we considered, including the NP-hard problem of MaxCut. Of key importance, adding GI enabled solving problems of hierarchical complexity, even when these problems were made to be asymmetric so as to avoid unfair comparisons due to the fact that GI has an bias-exploiting advantage on symmetric problems.

Adding GI to OM is elegant and did not introduce any further parameters. There are also no biases explicitly maintaining alternative patterns that have equal fitness that allow better performance on problems that have plateaus (inside hierarchical complexity). Although on problems with fully decomposable subfunctions, GI has drawbacks, in the light of modern dependency detection approaches, such problems should not be solved with EAs anymore. GI-GOMEA was found to perform better in terms of function evaluations required on all other problems that did have overlapping subfunctions. Interestingly, compared to GOMEA, the population size was reduced, sometimes by a factor of 10 (scaling to even larger factors for larger problems), even though GOMEA already typically requires much smaller population sizes than classical EDAs. These properties, we believe, make the combination of GI and OM a very interesting one to study further on the road to ever-more competent MBEAs, particularly for complex, multi-structured problems, and when combined with modern fitness-based linkage learning.


\newpage
\bibliographystyle{ACM-Reference-Format}
\bibliography{acmart}


\end{document}